\documentclass[10pt,twocolumn,letterpaper]{article}
\usepackage[english]{babel}
\usepackage{iccv}
\usepackage{times}
\usepackage{epsfig}
\usepackage{graphicx}
\usepackage{amsmath}
\usepackage{amssymb}
\usepackage[normalem]{ulem}
\usepackage{makecell}
\usepackage{courier}
\usepackage{bbm}
\usepackage[noend]{algpseudocode}
\usepackage{algorithmicx,algorithm}
\usepackage{adjustbox}
\usepackage{tabularx}
\usepackage{booktabs}
\usepackage{setspace}
\usepackage[font=small]{caption}
\algrenewcommand\algorithmicindent{1.0em}
\usepackage[pagebackref=true,breaklinks=true,letterpaper=true,colorlinks,bookmarks=false]{hyperref}

\def\c{\boldsymbol{\mathrm{c}}}

\def\x{\boldsymbol{\mathrm{x}}}
\def\v{\boldsymbol{\mathrm{v}}}
\def\q{\boldsymbol{q}}
\def\s{\boldsymbol{\mathrm{s}}}
\def\w{\boldsymbol{w}}

\def\X{\boldsymbol{\mathrm{X}}}

\def\y{\boldsymbol{\mathrm{y}}}
\def\z{\boldsymbol{\mathrm{z}}}
\def\W{\boldsymbol{\mathrm{W}}}
\def\w{\boldsymbol{\mathrm{w}}}
\def\U{\boldsymbol{\mathrm{U}}}
\def\V{\boldsymbol{\mathrm{V}}}
\def\P{\boldsymbol{\mathrm{P}}}

\def\Ee{\mathbb{E}}
\def\Re{\mathbb{R}}

\iccvfinalcopy 

\def\iccvPaperID{526} 

\ificcvfinal\fi
\begin{document}

\title{Structured Attentions for Visual Question Answering}

\author{Chen Zhu, Yanpeng Zhao, Shuaiyi Huang, Kewei Tu, Yi Ma\\
ShanghaiTech University\\
{\tt\small \{zhuchen, zhaoyp1, huangsy, tukw, mayi\}@shanghaitech.edu.cn}
}

\maketitle

\begin{abstract}
Visual attention, which assigns weights to image regions according to their relevance to a question, is considered as an indispensable part by most Visual Question Answering models. Although the questions may involve complex relations among multiple regions, few attention models can effectively encode such cross-region relations. In this paper, we demonstrate the importance of encoding such relations by showing the limited effective receptive field of ResNet on two datasets, and propose to model the visual attention as a multivariate distribution over a grid-structured Conditional Random Field on image regions. We demonstrate how to convert the iterative inference algorithms, Mean Field and Loopy Belief Propagation, as recurrent layers of an end-to-end neural network. We empirically evaluated our model on 3 datasets, in which it surpasses the best baseline model of the newly released CLEVR dataset \cite{johnson2016clevr} by 9.5\%, and the best published model on the VQA dataset \cite{antol2015vqa} by 1.25\%. Source code is available at \url{https://github.com/zhuchen03/vqa-sva}.
\end{abstract}

\section{Introduction}
Visual Question Answering (VQA) is a comprehensive task inspecting intelligent systems' ability to recognize images and natural languages together. Advances in this area not only benefit real-world applications which require the synergistic reasoning of vision and language, such as querying events in surveillance videos \cite{tu2014joint} and searching specific goods in images \cite{yeh2008photo}, but also call for finer-grained understanding on the semantic structures of images.
\begin{figure}
  \centering
  \includegraphics[width=0.9\linewidth]{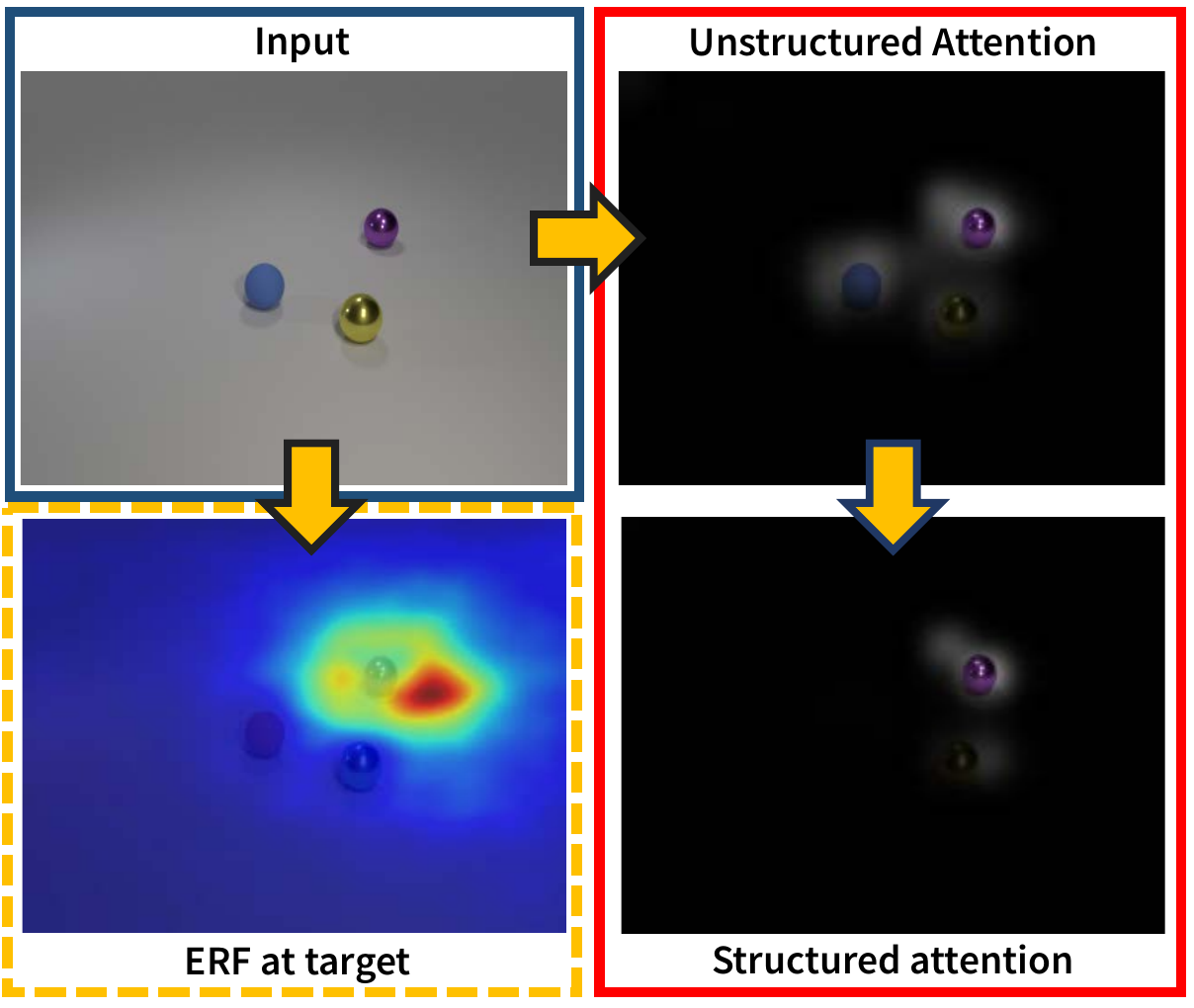}
  \caption{An example of the proposed model on CLEVR, demonstrating it is capable of inferring spatial relations despite the limited effective receptive field of the CNN. Question: \emph{What is the color of the sphere on the right of the metal sphere?} Our model overcomes the unstructured attention's tendency to attend to isolated key words in the questions, attending to the right region and giving the correct answer \emph{purple}. }
\label{fig:teaser}
  \vspace{-1em}
\end{figure}

Adoption of the visual attention mechanism like \cite{xu2016ask, yang2016stacked, lu2016hierarchical,nam2016dual} is a major source to boost  performance of VQA. Visual attentions impose regularization on the models to find the most relevant image regions to the question. Still, experiments \cite{das2016human} point out that state-of-the-art models often fail to identify the related regions like humans do. We argue this problem comes from the fact that such attention models do not take into account the spatial relations between regions when predicting the attention. This is important because the effective receptive field (ERF) of deep CNNs only covers a small fraction of the image \cite{luo2016understanding}. Even with memory mechanism \cite{xiong2016dynamic, nam2016dual,yang2016stacked,xu2016ask}, it is difficult to infer the right attention corresponding to questions that involve the spatial relations between regions without overlapping ERFs.

In this paper, we propose a novel neural network to model the attention with a multivariate distribution which considers the arrangements of image regions. We adopt the most straightforward graph structure to model image region arrangements - a grid-structured Conditional Random Field (CRF) \cite{nowozin2011structured}. The framework of the proposed method is illustrated in Fig. \ref{fig:framework}. We show that attention can be formulated as the marginal distribution of each hidden variable in the CRF. Then, we implement the iterative approximate inference algorithms, Mean Field and Loopy Belief Propagation, as recurrent layers of neural networks, which iteratively refines the attention. An example of this process is shown in Fig. \ref{fig:teaser}. We evaluate the proposed model on three representative datasets, where our model is competitive with the rule-based model \cite{andreas2016neural} on the SHAPES dataset, and surpasses the best baseline model of CLEVR by more than 9.5\% and best published method \cite{kim2016hadamard} on the VQA dataset by 1.25\%. We also demonstrate that the ERF on CLEVR and the VQA dataset is not large enough for previous methods to answer all questions involving object arrangements correctly.

Our work has the following contributions. First, we propose to model structured attentions with a CRF over image regions for Visual Question Answering, to address the problem of limited effective receptive fields of CNNs. Second, we demonstrate how to unfold both Mean Field and Loopy Belief Propagation algorithms for CRF as recurrent layers of neural networks, and perform comprehensive evaluation of the two different networks on three challenging datasets. Third, we give empirical evaluations of ERFs on CLEVR and the MSCOCO dataset.

\section{Related Work}
There have been many different directions for improving VQA performance, including predicting answer types \cite{kafle2016answer}, utilizing task-specific submodules \cite{andreas2016learning, andreas2016neural}, and better multimodal feature pooling methods \cite{fukui2016multimodal, kim2016hadamard}. Our focus is on structure-aware spatial attention applied on the visual feature. There are two major forms of spatial attention for VQA. One \cite{shih2016look, li2016visual, ilievski2016focused} is based on region proposals generated by Edge Box \cite{zitnick2014edge}, and the other \cite{xu2016ask, yang2016stacked, lu2016hierarchical, nam2016dual, fukui2016multimodal, kim2016hadamard} predicts attention on the individual feature vectors of convolutional feature maps. The most representative model SAN \cite{yang2016stacked} adopted multiple attention layers to support multi-step reasoning. There are also methods \cite{lu2016hierarchical, nam2016dual} that adopted both image attention and question attention to refine the image and question representation simultaneously. \cite{xu2016ask} proved its ability to recognize absolute and relative positions with two simple experiments, but the model itself does not consider the arrangement of regions and the success may be attributed to the power of CNNs.

Some of the methods considered the structures of images. \cite{malinowski2014multi} adopted the Bayesian framework based on the logic forms of segmentation results. The method was surpassed by some simple baseline models due to its demand for better semantic segmentation \cite{kafle2016visual}. The DMN \cite{xiong2016dynamic} adopted a bidirectional GRU that traverses the convolutional feature map in a snake-like fashion to encode the dependency of regions, which might not be the optimal choice for the 2D structure of images. \cite{li2016visual} concatenated the 8D bounding box representation with the 4096D visual feature, in which the spatial information could be easily overweighed by the visual feature.

There have been a number of approaches combining neural networks and CRF to predict structured outputs in both computer vision and natural language processing. \cite{peng2009conditional} utilized neural networks with sigmoid activation to predict the unary potential for sequential labeling. \cite{do2010neural} also used neural networks to predict the pairwise potential of labels. \cite{chen2015learning} learned fixed pairwise potentials for word recognition and image classification. \cite{jaderberg2014deep} utilized CNNs to predict both the unary potentials and higher-order potentials for unconstrained word recognition. \cite{zheng2015conditional} proposed to unfold the Mean Field algorithm as recurrent layers for semantic segmentation. It modeled the pairwise potential with Gaussian kernels, which encourages similar features to take the same label.  Besides the supervised structure in the output layers, \cite{kim2017structured} enforced the intermediate layers of neural networks to learn structured attentions for natural language tasks.

To the best of our knowledge, structured attention has not been explored for the complex task of VQA. We give the first empirical evaluations on unfolding both Mean Field and Loopy Belief Propagation as intermediate recurrent layers on the task of VQA, which can be seen as a further exploration of \cite{kim2017structured} in modeling 2D structures of visual data.

\section{Attention Models and Methods}
\begin{figure*}
  \centering
  \includegraphics[width=\linewidth]{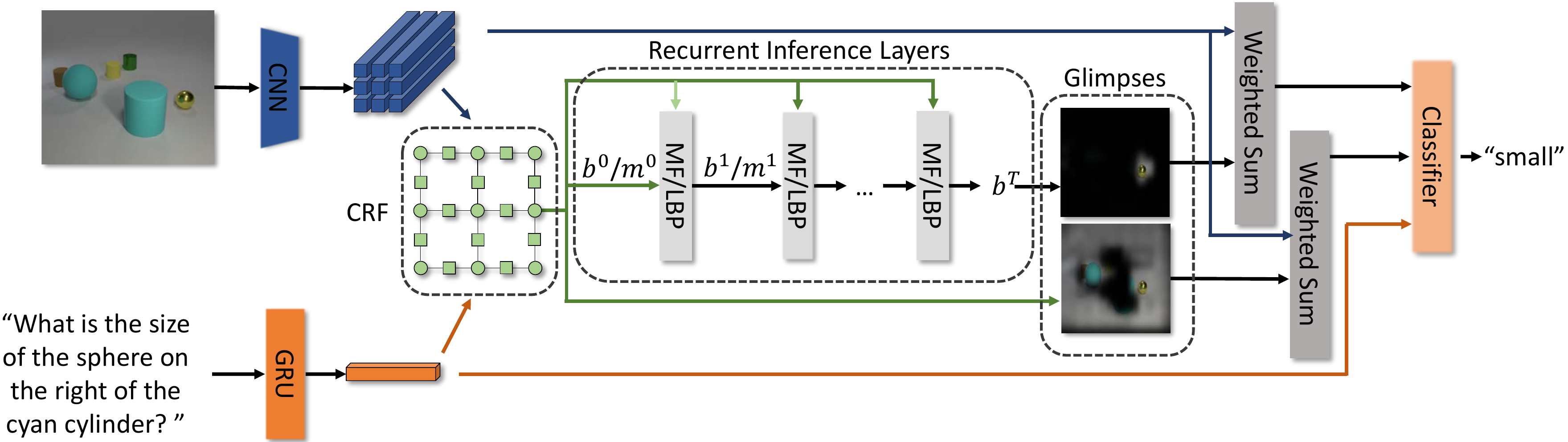}
  \caption{The whole picture of the proposed model. The inputs to the recurrent inference layers are the unary potential $\psi_{i}(z_i)$ and pairwise potential $\psi_{ij}(z_i,z_j)$, computed with Eq. \ref{eq:potentials}. $\psi_{i}(z_i)$ can also be used as an additional glimpse, which usually detects the key nouns. In the inference layers, $x^i$ represents $b^{(i)}$ for MF and $m^{(i)}$ for LBP. The recurrent inference layers generates a structured glimpse with MF or LBP. The 2 glimpses are used to weight-sum the visual feature vectors. The classifier use both of the attended visual features and the question feature to predict the answer. The demonstration is a real case. }\label{fig:framework}
  \vspace{-1.5em}
\end{figure*}
The general architecture of the proposed model is shown in Fig. \ref{fig:framework}. Here we define some notations used across the paper. We take the question feature $\q\in \Re^{n_Q}$ from the last time step of a GRU such as \cite{kiros2015skip}, and the image feature map $\X=[\x_{1},...,\x_{M}]\in\Re^{n_I\times M}$ from one of the convolution layers of a CNN such as \cite{he2016deep}. Here $n_Q, n_I$ are the dimensions of question and image feature vectors respectively, and $M$ is the total number of image feature vectors which divides the image into $M$ regions. We use $\mathrm{softmax}(\cdot)$ to denote the softmax activation function and $\sigma(\cdot)$ to denote the sigmoid activation function. The attention mechanism in VQA aims to produce a \emph{context} $\c$ from $\X$ which represents the visual feature related to the question.
\subsection{Unstructured Categorical Attention}
In previous methods for VQA, visual attention is usually modeled as a single or multi-step soft-selection from $\X$. As shown in \cite{xu2015show,kim2017structured}, the soft-selection approach represents the selected region index by a categorical latent variable $z\in\{1,...,M\}$ and defines $\c$ as an expectation of the selection:
\begin{equation}\label{eq:coarse_context}
   \c = \Ee\left[\sum_{i}\mathbbm{1}_{\{z=i\}}\cdot\x_i\right]=\sum_{i}p(z=i|\X,\q)\x_i,
\end{equation}
where $\mathbbm{1}_{\{z=i\}}$ is an indicator function, and the distribution of $z$ is parameterized by
\begin{equation}\label{eq:attention}
   p(z=i|\X,\q) = \mathrm{softmax}(\U g(\x_i,\q)),
\end{equation}
where $\U\in\Re^{1\times n_I}$ and $g(\cdot)$ is some multimodal feature pooling function such as \cite{fukui2016multimodal, kim2016hadamard}. Noticing this model ignores the spatial arrangement of the feature vectors in $\X$ in each step, and the resulting hidden states in multi-step models \cite{xu2015show, xu2016ask} are still unstructured, we have dropped the hidden states in the condition presented in \cite{xu2016ask} for notational convenience.

Since categorical distribution only requires the probabilities to be positive and sum to 1, the following normalized sigmoid attention is still a valid categorical attention:
\begin{equation}\label{eq:sigmoid_attn}
   p(z=i|\X,\q)=\frac{\sigma(\U g(\x_i,\q))}{\sum_{j}\sigma(\U g(\x_j,\q))}.
\end{equation}
We can use such attention as a glimpse\footnote{Glimpses refer to multiple attentions, same as in \cite{fukui2016multimodal, kim2016hadamard}.} of our model, which will be introduced in Section \ref{sec:concat}.

\subsection{Structured Multivariate Attention}
To consider the arrangement of $\X$, we adopt a structured multivariate attention model similar to \cite{kim2017structured}, in which we consider the distribution $\z\sim p(\z|\X,\q)$ as a vector of binary latent variables $\z=[z_{1},...,z_{M}]^T$ with $z_{i}=1$ if $\x_{i}$ is related to the question and $z_{i}=0$ otherwise. Multiple regions can now be selected at the same time. We define the context as the expectation of the sum over all related regions, which can be derived as a sum of $\x_i$ weighted by the marginal probability $p(z_{i}=1|\X,\q)$:
\begin{equation}\label{eq:attn}
  \Ee_{\z\sim p(\z|\X,\q)}[\X\z]=\sum_{i}p(z_{i}=1|\X,\q)\x_{i}.
\end{equation}
Let $S=\sum_{i}p(z_{i}=1|\X, \q)$. Since $0\le S\le M$ and $M$ is relatively large, to reduce covariate shift, we normalize the expectation to get the context $\c$:
\begin{equation}\label{eq:context}
  \c = \frac{1}{S}\sum_{i}p(z_{i}=1|\X,\q)\x_{i},
\end{equation}

We model the distribution $p(\z|\X,\q)$ in the most straightforward form, a grid-structured Conditional Random Field, which represents the joint probability $p(\z|\X,\q)$ with a grid-structured factor graph that considers the pairwise joint distribution of a region's 4-neighbourhood, as shown in Fig. \ref{fig:framework}. Let $\mathcal{N}=\{(i,j)|i<j, j\in N_i\}$, where $N_i$ is the set of $i$'s neighbors on the graph. The grid-structured CRF assumes
\begin{equation}
\label{eq:gridcrf}
  p(\z|\X,\q) = \frac{1}{Z}\prod_{(i,j)\in \mathcal{N}}\psi_{ij}(z_i,z_j)\prod_{i}\psi_{i}(z_i),
\end{equation}
where the unary potential $\psi_{i}(z_i)\ge 0$ measures the likelihood of region $i$ taking the value $z_i\in \{0,1\}$, and the pairwise potential $\psi_{ij}(z_i,z_j)\ge 0$ measures the likelihood of regions $(i, j)$ taking values $z_i, z_j$ respectively.

\subsection{Recurrent Inference Layers}
The inference problem in such a gird-structured factor graph, which aims to calculate the marginal probability $p(z_i|\X,\q;\theta)$, is known to be NP-hard \cite{shimony1994finding}. Still, there are approximate inference algorithms to solve the problem, such as Mean Field (MF) and Loopy Belief Propagation (LBP). These algorithms take potential functions $\psi_{i}(z_i)$ and $\psi_{ij}(z_i,z_j)$ as inputs and update $p(z_i|\X,\q;\theta)$ iteratively through message passing.  We train neural networks to predict optimal $\psi_{i}(z_i)$ and $\psi_{ij}(z_i,z_j)$ and then run the algorithms for a fixed number of $T$ steps. The iterative algorithms are implemented as recurrent inference layers in the neural network.

\vspace{-0.6em}
\subsubsection{Potential Functions}
In VQA, the potential functions should depend on both the image and the question. We use low-rank bilinear pooling \cite{kim2016hadamard}, a parsimonious bilinear model, to capturing the interaction between 2 feature vectors $\x, \y$:
\begin{equation}\label{eq:mlb}
  g(\x,\y;\P_x,\P_y) = \mathrm{tanh}(\P_x\x)\odot \mathrm{tanh}(\P_y\y),
\end{equation}
where $\P_x,\P_y$ are learnt projection matrices projecting $\x,\y$ to the same dimension, and $\odot$ represents Hadamard product.
Based on this, we model $\psi_{i}(z_i)$ and $\psi_{ij}(z_{i},z_{j})$ as following:
\begin{equation}\label{eq:potentials}
  \psi_{i}(z_i=1) = \sigma\left(\U g\left(\x_i, \q;\U_x,\U_q\right)\right),
\end{equation}
\begin{equation}
  \psi_{i}(z_i=0) = 1-\psi_{i}(z_i=1),
\end{equation}
\begin{equation}
  \psi_{ij}(z_i, z_j)  =h\left(\v_{z_iz_j} g\left(\y_{ij}, \q;\V_y, \V_q\right)\right),
\end{equation}
where $\U_x\in \Re^{n_c\times n_I}, \U_q\in\Re^{n_c\times n_Q},~ \U\in\Re^{1\times n_c}, \V_{y}\in\Re^{n_c\times 2n_I}, \V_q\in\Re^{n_c\times n_Q}$ are learnt projection matrices, $\v_{z_iz_j}$ is a row vector of $\V\in\Re^{4\times n_c}$ indexed by $z_i$ and $z_j$, $\y_{ij}=[\x_i^T, \x_j^T]^T$, $n_c$ is the common projection dimension, $h(\cdot)$ is a certain activation function which will differ in the 2 inference algorithms.

\vspace{-0.8em}
\subsubsection{Mean Field Layers}
\begin{algorithm}[t]\small
\caption{\small MF Recurrent Layer in VQA}
\label{alg:MF}
\hspace*{0.02in} {\bf Input:}
$\psi_{i}(z_i)$, $\ln \psi_{ij}(z_i, z_j)$
\begin{algorithmic}[0]
\State initialize $b_i^{(0)}(z_i) = \psi_{i}(z_i)$
\For{t=1:T}
\For{$i=1:M$, $z_i$ in $\{0,1\}$}
   \State s$\leftarrow$0
   \For{$j$ in $N_i$, $z_j$ in $\{0,1\}$}
   \State s$\leftarrow$s$+b_j^{(t-1)}(z_j)\ln\psi_{ij}(z_i,z_j)$
   \EndFor
   \State $b_i^{(t)}(z_i)\leftarrow \psi_{i}(z_i)\exp(s)$
   \State normalize($b_i^{(t)}$)
\EndFor
\EndFor
\State \Return $b_i^{(T)}(z_i)$
\end{algorithmic}
\end{algorithm}

The Mean Field algorithm approximates the distribution $p(\z|\X,\q)$ in Eq. \ref{eq:gridcrf} with a fully factorized distribution $q(\z)$:
\begin{equation*}
  q(\z) = \prod_{i}b_{i}(z_i),
\end{equation*}
where $b_i(z_i)$ are variational parameters corresponding to the marginal probabilities $p(z_i|\X, \q)$. The variational parameters are optimized by iteratively minimizing the mean-field free energy
\begin{small}
\begin{equation}\label{eq:mf_energy}
\begin{split}
   F_{MF}({b_i}) =& -\sum_{(i,j)\in \mathcal{N}}\sum_{z_i,z_j}b_i(z_i)b_j(z_j)\ln\psi_{ij}(z_i,z_j) \\
     & +\sum_{i}\sum_{z_i}b_i(z_i)[\ln b_i(z_i)-\ln \psi_i(z_i)]
\end{split}
\end{equation}
\end{small}
\hspace{-0.3em}subject to the constraint $\sum_{z_i}b_i(z_i)=1$, which is shown to be equivalent to minimizing the KL divergence between $p(\z|\X,\q)$ and $q(\z)$ \cite{weiss2001comparing}. Specifically, MF initializes $b^{(0)}_i(z_i) = \psi_{i}(z_i)$ and updates $b_i^{(t)}(z_i)$ as:
\begin{small}
\begin{equation}\label{eq:mf_iter}
b_i^{(t)}(z_i)=\alpha\psi_{i}(z_i) \cdot\exp\left(\sum_{j\in N_i}\sum_{z_j}b_j^{(t-1)}(z_j)\ln\psi_{ij}(z_i,z_j)\right),
\end{equation}
\end{small}
\hspace{-0.3em}where $\alpha$ is the normalizing constant. Since Eq. \ref{eq:mf_iter} only involves $\ln\psi_{ij}(z_i,z_j)$, we adopt a log model for pairwise potential,
\begin{equation}
  \ln\psi_{ij}(z_i, z_j) =\mathrm{tanh}(\v_{z_i,z_j} g(\y_{ij}, \q; \V_y,\V_q)).
\end{equation}
MF can be unfolded as a recurrent layer of neural networks without parameters, where in each step $t$ the inputs are $\psi_i(z_i), \psi_{ij}(z_i,z_j)$ and $b_i^{(t-1)}(z_i)$, as demonstrated by Algorithm \ref{alg:MF}.

\vspace{-0.6em}
\subsubsection{Loopy Belief Propagation Layers}

\begin{algorithm}[t]\small
\caption{\small LBP Recurrent Layer in VQA}
\label{alg:LBP}
\hspace*{0.02in}{\bf Input:}
$\psi_{i}(z_i)$, $\psi_{ij}(z_i, z_j)$
\begin{algorithmic}[0]
\State initialize $m_{ij}^{(0)}(z_j)=0.5$, $m_{ij}^{(t)}(z_j)=0$ for $t>0$
\For{t=1:T}
\For{$j=1:M$, $i$ in $N_j$}
   \For{$z_i$ in $\{0,1\}$}
     \State s$(z_i)\leftarrow$1
     \For{ $k$ in $N_i\setminus\{j\}$}
        \State s$(z_i)\leftarrow$ s$(z_i)\cdot m_{ki}^{(t-1)}(z_i)$
     \EndFor
   \EndFor
     \For{$(z_i,z_j)$ in $\{0,1\}\times\{0,1\}$}
        \State $m_{ij}^{(t)}(z_j)\leftarrow m_{ij}^{(t)}(z_j)+\psi_{ij}(z_i,z_j)\psi_i(z_i)s(z_i)$
     \EndFor
   \State normalize($m_{ij}^{(t)}$)
\EndFor
\EndFor
\For{$i=1:M$}
   \State $b_i(z_i)\leftarrow \psi_i(z_i)$
   \For{$k$ in $N_i$, $z_i$ in $\{0,1\}$}
      \State $b_i(z_i)\leftarrow b_i(z_i)\cdot m_{ki}^{(T)}(z_i)$
   \EndFor
   \State normalize($b_i$)
\EndFor
\State \Return $b_i(z_i)$
\end{algorithmic}
\end{algorithm}

Loopy Belief Propagation can be applied similarly by initializing all messages to a fixed value and updating the messages iteratively in a fixed or random order \cite{nowozin2011structured}. In our grid-structured graph, since each factor is connected to only 2 variables, the variable-to-factor and factor-to-variable messages can be merged as a single message $m_{ij}$, representing the message from $z_i$ to $z_j$. We initialize $m_{ij}^{(0)}(z_j)=0.5$, and iteratively update the messages based on the formula
\begin{small}
\begin{equation}\label{eq:lbp_update}
m_{ij}^{(t)}(z_j)=\alpha\sum_{z_i}\psi_{ij}(z_i,z_j)\psi_{i}(z_i)
  \cdot \prod_{k\in N_i\setminus \{j\}}m_{ki}^{(t-1)}(z_i),
\end{equation}
\end{small}
\hspace{-0.3em}where $\alpha$ is the normalizing constant as above.
After a fixed number of steps $T$, the variable $z_i$ gathers all the messages from its neighbourhood to get the marginal probability:
\begin{equation}\label{eq:lbp_marginal}
  b_i(z_i)=\beta\psi_{i}(z_i)\prod_{k\in N_i}m^{(T)}_{ki}(z_i),
\end{equation}
where $\beta$ is the normalizing constant.

Similar to MF, LBP can also be unfolded into parameterless recurrent layer, where at each time step $t$ the input is $\psi_i(z_i), \psi_{ij}(z_i,z_j)$ and $m^{(t-1)}_{ij}(z_j)$, and after $T$ steps $m^{(T)}(z_i)$ is used to compute marginal probability $b_i(z_i)$, as shown in Algorithm \ref{alg:LBP}.

\subsection{The overall structure}\label{sec:concat}
As shown in Fig. \ref{fig:framework}, the overall structure of the proposed model is an end-to-end classification neural network. Firstly, $\psi_{i}(z_i)$ and $\psi_{ij}(z_i,z_j)$ are computed using the extracted features $\X$ and $\q$. Then, the recurrent inference layers run for $T$ steps to get the structured marginal probability $p(z_i=1|\X,\q)=b_i^{(T)}(z_i=1)$, which is then used to compute the structured context $\hat{\c}$ with Eq. \ref{eq:context}. We also compute a unstructured context $\tilde{\c}$ with Eq. \ref{eq:coarse_context} by replacing $\U g(\x_i,\q)$ in Eq. \ref{eq:sigmoid_attn} with $\psi_i(z_i=1)$. In the classifier, the contexts are both pooled with the question to get
\begin{equation}\label{eq:context_ans}
  \hat{\s} = g(\hat{\c},\q; \hat{\W}_{c},\hat{\W}_{q}),
\end{equation}
\begin{equation}
  \tilde{\s} = g(\tilde{\c}, \q; \tilde{\W}_{c},\tilde{\W}_q),
\end{equation}
where $\hat{\W}_{c},\tilde{\W}_{c}\in \Re^{n_c\times n_I}$, $\hat{\W}_{q},\tilde{\W}_q\in\Re^{n_c\times n_Q}$. The answer is predicted with
\begin{equation}\label{eq:answer}
  a = \arg\max_{k\in\Omega_K} \mathrm{softmax}(\w_k[\hat{\s}^T, \tilde{\s}^T]^T),
\end{equation}
where $\w_k$ is the $k$-th row of $\W\in\Re^{K\times 2n_c}$, $K$ is the number of answers, $\Omega_K$ is the answer space with up to $K$ answers.

\section{Experiments}
\subsection{Datasets}
\textbf{The SHAPES dataset} \cite{andreas2016neural} is a synthetic dataset consisting of images containing 3 basic shapes in 3 different colors with a resolution of 30$\times$30, and queries about the arrangements of the basic shapes, as shown in Fig. \ref{fig:shapes_attention}. The answer is ``yes" when the image satisfies the query, and ``no" otherwise. There are 3 different lengths of queries. The original dataset \cite{andreas2016learning} has 14592 and 1024 image/question pairs for the training and test sets. All the queries in the test set do not appear in the training set.

\textbf{The CLEVR dataset} \cite{johnson2016clevr} is a much more complex but unbiased synthetic dataset aiming at testing visual abilities such as counting, comparing and logical reasoning. It consists of 100,000 images of simple 3D objects with random shapes, sizes, materials, colors and positions with a resolution of 320$\times$480, and nearly a million natural language questions, 853,554 of which are unique. The questions can be categorized into 5 general types: \emph{exist}, \emph{count}, \emph{compare integer}, \emph{query attribute}, and \emph{compare attribute}. There are 699,989 training questions, 149,991 validation questions and 149,988 test questions. The vocabulary sizes for the questions and answers are 82 and 28 respectively.

\textbf{The VQA real-image dataset} \cite{antol2015vqa} is a comprehensive dataset which requires knowledge beyond the dataset to answer all the questions correctly. It has about 204,721 images from MSCOCO \cite{lin2014microsoft} each with 3 natural language questions, and each question has 10 answers collected from online workers. It consists of 3 splits: \texttt{train}, \texttt{val} and \texttt{test}, each of which has 248,349, 121,512 and 244,302 questions respectively. \texttt{test-dev} is a subset of \texttt{test}, which has 60,864 questions. We keep a collection of 2,000 most frequent answers from the union of \texttt{train} and \texttt{val}, and ignore questions with no answers from this collection, which leaves us 334,554 samples for training. With the same preprocessing procedure as \cite{fukui2016multimodal, kim2016hadamard}, we also get 837,298 training samples from Visual Genome \cite{krishna2016visual} for augmentation.

\subsection{Model Configuration and Training}
For extracting image features, we use a 2-layer LeNet trained with the whole network on SHAPES as in \cite{andreas2016learning}, and ImageNet-pretrained ResNets \cite{he2016deep} on CLEVR and the VQA dataset. For sentence embedding, we use single-layer GRU. On the SHAPES dataset, we set $n_c=128, n_Q=128, n_I=50, M=9$. On CLEVR, following \cite{johnson2016clevr}, we resize the input images to $224\times 224$, and use feature maps at the \texttt{res4b22} layer of ResNet-101 ($n_I=1024, M=196$) and the \texttt{res5c} layer of ResNet-152 ($n_I=2048, M=49$) in different experiments. We also set $n_Q=2048$, comparable to the 2-layer LSTM with 1024 units per layer used in \cite{johnson2016clevr}. On the VQA dataset, we fix $n_c=1200$. Images are resized to $448\times 448$ and we use the feature at the \texttt{res5c} layer of ResNet-152 ($n_I=2048, M=196$ ), the same as \cite{lu2016hierarchical, fukui2016multimodal, kim2016hadamard}. We set $n_Q=2400$ since we use the pre-trained skip-thought vector \cite{kiros2015skip} provided by \cite{kim2016hadamard} as initialization.

For training, we implement our network with MXNet \cite{chen2015mxnet}. In all 3 tasks, we use the Adam optimizer \cite{kingma2014adam} with the default setting except for the learning rate, which is picked using grid search. We adopt Bayesian dropout \cite{gal2016theoretically} for GRU's as in \cite{kim2016hadamard}, and apply dropout \cite{srivastava2014dropout} before every other fully connected layer. On SHAPES and CLEVR, we find setting both dropout probability to 0.2 to be optimal, while on the VQA dataset, setting a small Bayesian dropout 0.25 for the GRU and a large dropout of 0.5 for the other parts achieved the optimal results. In addition, we use answer sampling by default on the VQA dataset as in \cite{fukui2016multimodal, kim2016hadamard}.

\subsection{Visualization of ERF}
To visualize the ERF, we need to compute the influence of a pixel $I_{ij}\in \Re^3$ on the entry $y_{rc}^n$, the feature at $(r,c)$ of the $n$-th channel in a certain conv layer, represented by $\Vert\partial y_{rc}^n/\partial I_{ij}\Vert$. As in \cite{luo2016understanding}, we assume a loss function $l_{rc}^n$ which is related only to channel $y_{rc}^n$, i.e., $\partial l_{rc}^n/\partial y_{rc}^n=1$ and $\partial l_{rc}^n/\partial y_{ij}^n=0$ for $i\neq r$ or $c\neq j$, so that $\partial y_{rc}^n/\partial I_{ij} = \partial l_{rc}^n/\partial I_{ij}$, since $\partial l_{rc}^n/\partial I_{ij}$ can be computed efficiently by DL frameworks. Finally, we draw the heat map of the total effect of a subset $\Omega_C$ of channels:
\begin{small}
\begin{equation}\label{eq:erf}
  E_{ij} = \left\Vert\sum_{n\in \Omega_C} \left(\frac{\partial l_{rc}^n}{\partial I_{ij}}\right)^2\right\Vert.
\end{equation}
\end{small}

\subsection{Results and Analysis}
We perform experiments on the 3 datasets, in which we fix the general structure except for the visual attention models, to quantify the role of structured attention in our model and the best configuration for it. We will use the following abbreviations to distinguish the models we implemented:
\vspace{-0.5em}
\begin{itemize}
\setlength\itemsep{0.1em}
  \item SM/SIG: 1-glimpse softmax or sigmoid attention.
  \item MF/LBP: 1-glimpse multivariate attention with a MF or LBP recurrent layer and a default $T=3$.
  \item \emph{MODEL}-G2: 2-glimpses with a default $T=3$, where \emph{MODEL}=SIG,MF or LBP is the attention model.
  \item MF-SIG/LBP-SIG: 2-glimpse model by concatenating a MF or LBP attention with a SIG attention, as mentioned in Section \ref{sec:concat}.
  \item \emph{MODEL}-T$n$: \emph{MODEL} with $T=n$ inference steps.
\end{itemize}

\subsubsection{On SHAPES}
\begin{figure}
  \centering
  \includegraphics[width=0.8\linewidth]{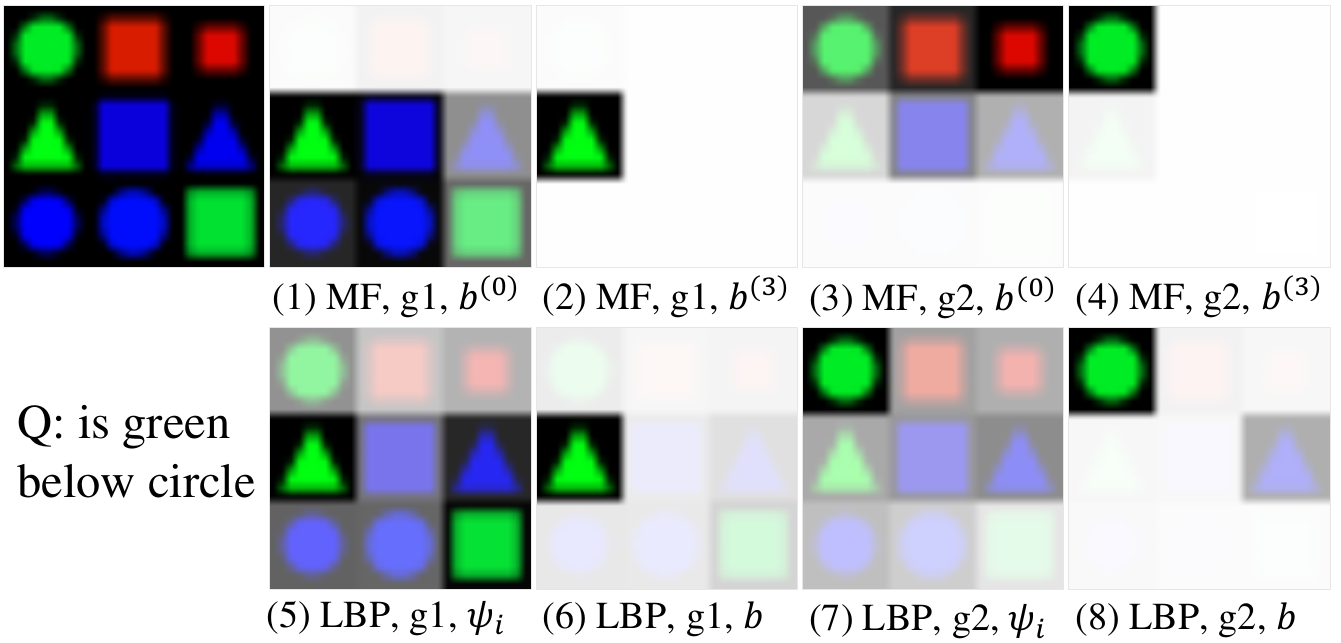}
  \caption{Visualization of MF-G2 and LBP-G2 on the test set of \texttt{large}. $b^{(0)}$ or $\psi_i$ represents the initial attention, $b^{(3)}$ or $b$ represents the refined attention after MF or LBP. The query looks for a green object under a circle.}\label{fig:shapes_attention}
  \vspace{0em}
\end{figure}

\begin{table}
\begin{center}
\makebox[\linewidth]{\resizebox{0.8\linewidth}{!}{%
  \begin{tabular}{llcccc}
  \Xhline{2\arrayrulewidth}
  & Query Length &  3 &  4 &  5 & All  \\
  & \% of test set & 12.5 & 62.5 & 25 & -   \\
  \hline
  \texttt{small} & NMN\cite{andreas2016neural} & \textbf{91.4} & \textbf{95.6} & \textbf{92.6} & \textbf{94.3} \\
   & SIG-G2 & 57.0 & 70.5 & 66.8 & 67.9 \\
   & MF-G2  & 53.1 & 71.4 & 66.0 & 67.8 \\
   & LBP-G2 & 63.3 & 72.2 & 62.5 & 68.7 \\
  \hline
   \texttt{medium} & NMN & \textbf{99.2} & 92.1 & \textbf{85.2} & 91.3 \\
   & SIG-G2 & 68.8 & 79.6 & 73.8 & 76.8 \\
   & MF-G2 & 98.0 & \textbf{99.6} & 71.5 & \textbf{92.4} \\
   & LBP-G2& 87.1 & 99.5 & 71.9 & 91.0 \\
  \hline
  \texttt{large} & NMN & \textbf{99.7} & 94.2 & \textbf{91.2} & 94.1 \\
  & SIG-G2 & 93.2 & 95.6 & 72.5 & 89.5 \\
  & MF-G2  & \textbf{99.7} & 99.9 & 79.2 & \textbf{94.7} \\
  & LBP-G2 & 95.1 & \textbf{100} & 78.9 & 94.1 \\
  \Xhline{2\arrayrulewidth}
\end{tabular}}}
\vspace{.1em}
\caption{Accuracy on SHAPES.}\label{tb:shapes}
\vspace{-1.5em}
\end{center}
\end{table}

In this part, we will look into the influence of the volume of the data on our model's generalization, and compare the performance of SIG-G2, MF-G2 and LBP-G2 models. We find it is difficult for all 3 models to generalize with the same amount of data as \cite{andreas2016learning}. This may because \cite{andreas2016learning} uses a parser to understand the queries with guaranteed correctness on this dataset, while we have to train the GRU to understand the queries from scratch. The parser-based method may not perform well in more general tasks such as the VQA dataset, since they found using fewer modules on the VQA dataset turned out to be better, but our RNN-based approach should generalize better with enough training data. So we generate more data with the same answer distributions for each query as \cite{andreas2016learning} to train and test our model, and re-trained their model using the released code. We name the original dataset \texttt{small}, and the newly generated datasets with 2 and 3 times as much data in both training and test sets as \texttt{medium} and \texttt{large} respectively. We find with more training data, our model becomes competitive with \cite{andreas2016learning} and the MF-G2 model surpasses it on both \texttt{medium} and \texttt{large}, as shown in Table \ref{tb:shapes}. Our models are extremely good at handling length-4 queries, which looks for object arrangements in 4-neighborhood, as demonstrated in Fig. \ref{fig:shapes_attention}. The high accuracy also implies the model is capable of set theory reasoning, since it achieved high test accuracy with length-3 queries which contain self-conflict queries such as \emph{is red green}, which aims to find an object that is both red and green. For complex queries, such as \emph{is red below below green}, which aims to find a red object below another object that is below a green object, it is not as competitive as \cite{andreas2016learning}, probably because the GRU in our model has not generalized to higher order logic.

\subsubsection{On CLEVR}
\begin{figure}
\begin{minipage}[t]{0.33\linewidth}
\centering
\includegraphics[width=0.9\linewidth]{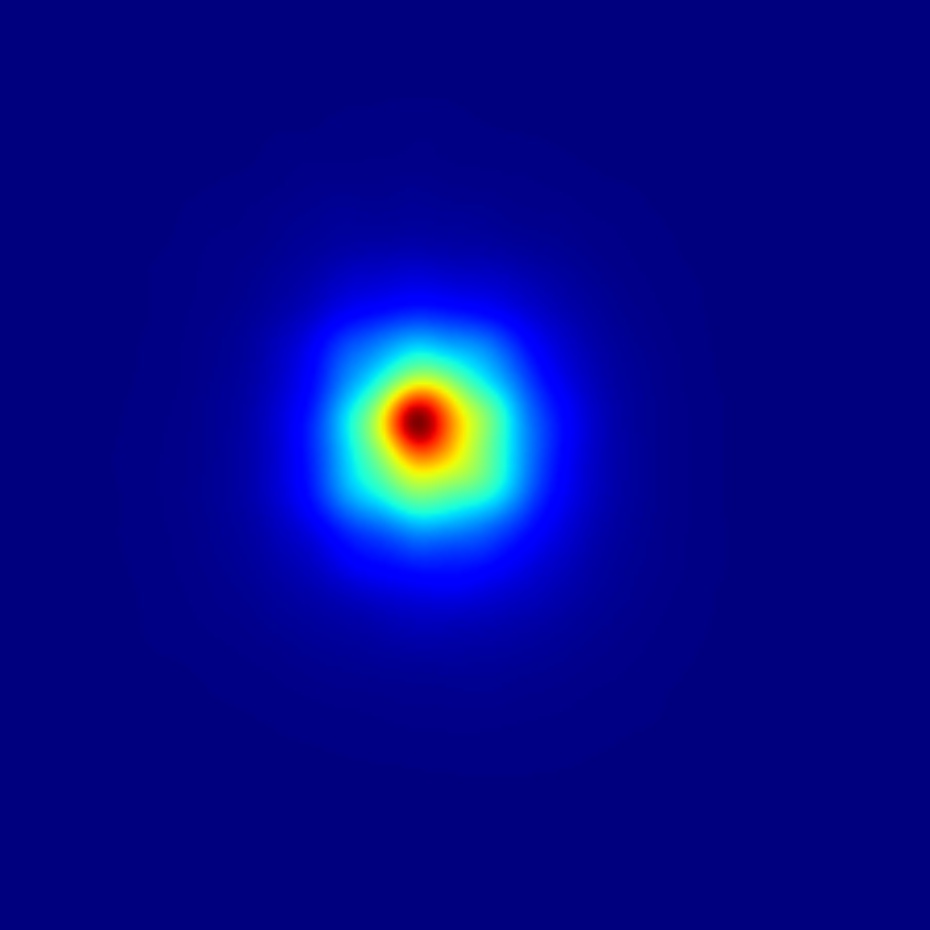}
\caption*{\begin{small}\texttt{res5c} on CLEVR\end{small}}
\end{minipage}%
\begin{minipage}[t]{0.33\linewidth}
\centering
\includegraphics[width=0.9\linewidth]{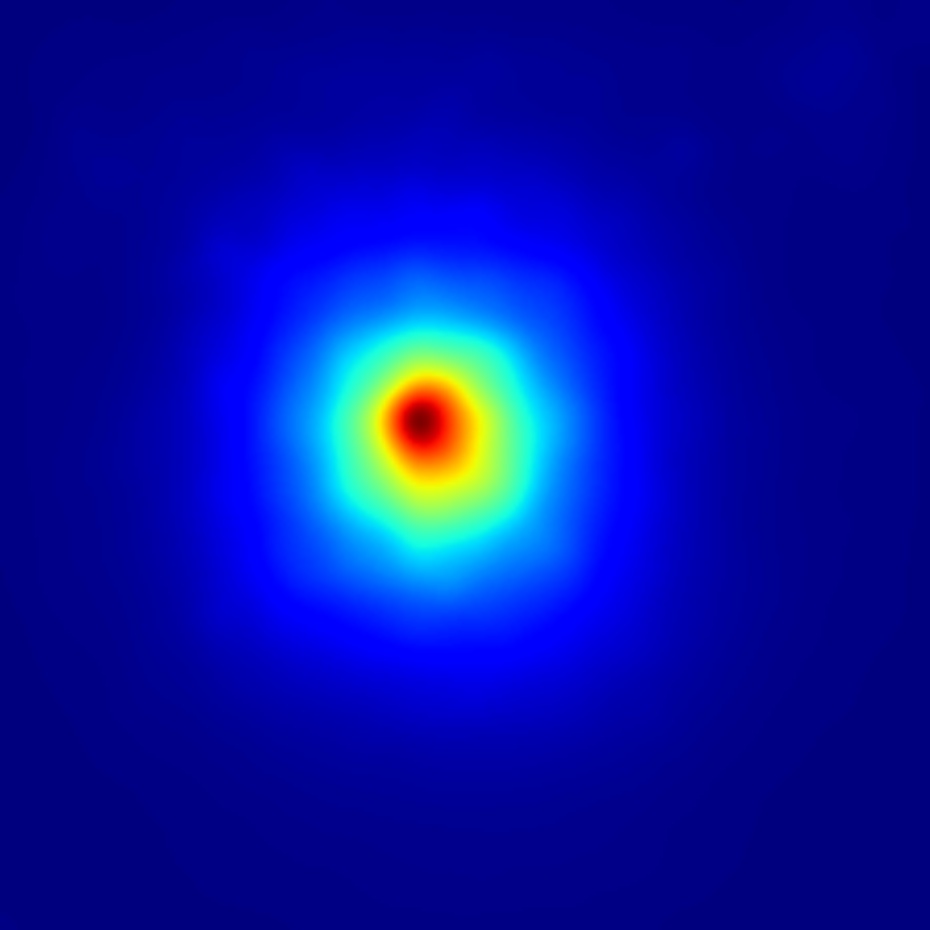}
\caption*{\begin{small}\texttt{res5c} on CLEVR\end{small} }
\end{minipage}
\begin{minipage}[t]{0.33\linewidth}
\centering
\includegraphics[width=0.9\linewidth]{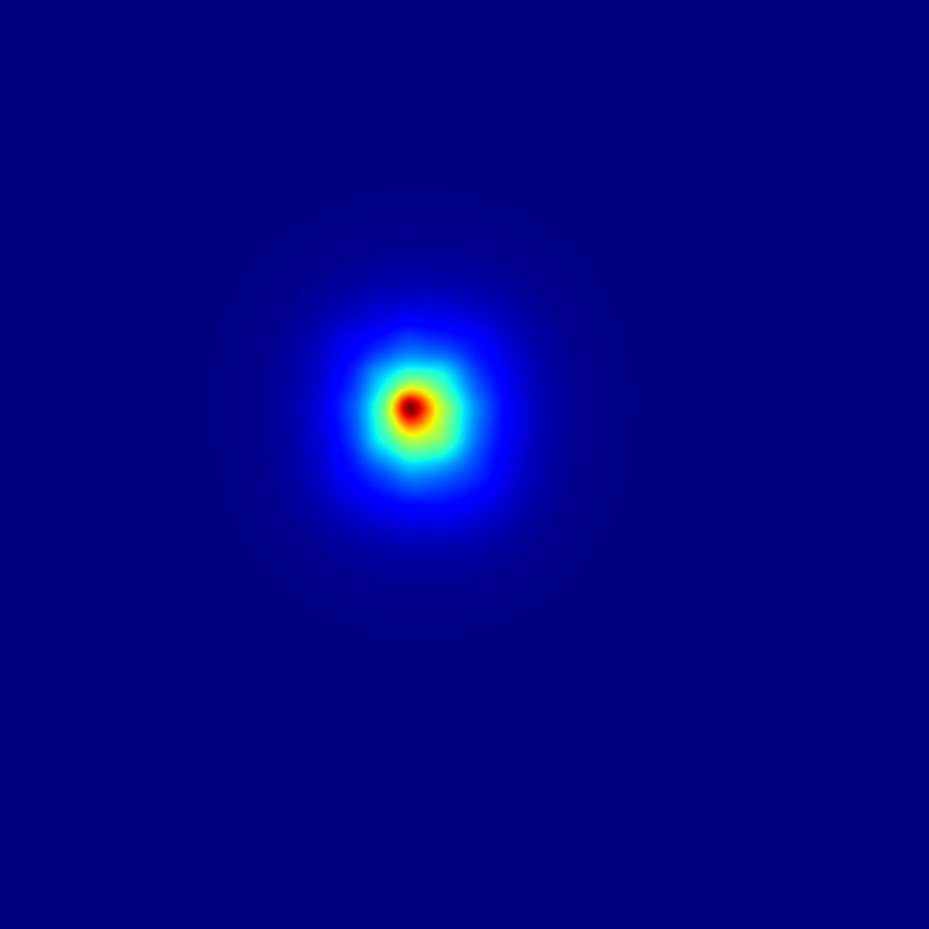}
{\caption*{\begin{small}\texttt{res5c} on MSCOCO\end{small} }}
\end{minipage}
\caption{The average ERF \cite{luo2016understanding} of 32 channels chosen at regular intervals, on 15000 images from the CLEVR test set and 52500 images from the MSCOCO test set with resolution of $224\times 224$ and $448\times 448$ respectively. The ERF images are smoothed with $\sigma=4$ Gaussian kernels.}\label{fig:clevr_erf}\vspace{-1em}
\end{figure}

In this part, we study the role of different visual features and different kinds of attentions on the performance, and test our best models on the test set, as shown in Table \ref{tb:clevr}. Our best model surpasses the best baseline model in \cite{johnson2016clevr} by more than 9.5\% on the test set. Both MF and LBP outperform SM and SIG, demonstrating the effectiveness of our method. The maximum margin of MF/LBP vs. SIG on overall accuracy is 2.62\% and 1.33\% with the ResNet-152 and ResNet-101 features respectively. The most significant improvement of MF/LBP over other models is on Compare Attribute, which involves comparing specific attributes of objects specified by spatial relations with other objects. This also proves that our model alleviates the problem of previous methods of ignoring arrangements of regions. Further, we show the receptive fields of the 2 selected layers on the test set in Fig. \ref{fig:clevr_erf}. In each image, we choose the feature vector closest to the center, where there is a higher chance for objects to appear. Still, both of them only occupy a small portion of the image, indicating the importance of considering the structure of regions. Overall, the performance with \texttt{res4b22} features is better than that with \texttt{res5c} features. From the ERF point of view, the ERF of \texttt{res5c} has more than twice the area as the ERF of \texttt{res4b22}. As a result, the feature vector of \texttt{res5c} may require more than twice the number of parameters to represent same amount of information in this region as \texttt{res4b22}, but its has only twice.

\begin{table}
\begin{center}
\makebox[\linewidth]{\resizebox{0.9\linewidth}{!}{%
		\begin{tabular}{lcccccc}
			\Xhline{2\arrayrulewidth}
            & All & Exist & Count & CI & QA & CA \\
            \cmidrule{2-7}
            \texttt{res5c} & \multicolumn{6}{c}{CLEVR validation}\\
            \hline
			SM & 68.80 & 73.20 & 53.16 & 76.52 & 81.58 & 56.77 \\
			SIG & 70.52 & 73.90 & 53.89 & 76.52 & 82.46 & 63.06 \\
			MF & 73.14 & 76.46 & \textbf{56.89} & 77.43 & 83.72 & \textbf{68.76} \\
			LBP & 72.30 & 76.32 & 54.92 & 77.50 & 83.35 & 67.54 \\
			MF-SIG & 73.19 & 76.53 & 56.22 & \textbf{78.56} & \textbf{84.23} & 68.34  \\
			LBP-SIG & \textbf{73.33} & \textbf{77.50} & 56.39 & 77.97 & 84.09 & 68.70 \\
			\cmidrule{1-7}
			\texttt{res4b22} &  \multicolumn{6}{c}{CLEVR validation}\\
             \hline
			SM & 75.63 & 77.69 & 57.79 & 78.63 & 87.76 & 71.83 \\
			SIG & 75.32 & 76.54 & 58.93 & 78.12 & 87.94 & 69.38 \\
			MF & 76.65 & 77.90 & 58.87 & \textbf{80.48} & 88.10 & 74.34 \\
			LBP & 76.21 & 78.97 & 57.52 & 80.14 & 87.90 & 73.43 \\
			MF-SIG& 77.4 & \textbf{79.8} & 61.0 & 79.3 & 88.0 & 75.1  \\
			LBP-SIG & \textbf{77.97} & 79.7 & \textbf{61.39} & 80.17 & \textbf{88.54} & \textbf{76.31} \\
			\cmidrule{1-7}
             \texttt{res4b22} &  \multicolumn{6}{c}{CLEVR test}\\
             \hline
			MCB\cite{johnson2016clevr} & 51.4 & 63.4 & 42.1 & 63.3 & 49.0 & 60.0  \\
			SAN\cite{johnson2016clevr} & 68.5 & 71.1 & 52.2 & 73.5 & 85.2 & 52.2  \\
			MF-SIG& 77.57 & \textbf{80.05} & 60.69 & 80.08 & 88.16 & 75.27  \\
			LBP-SIG & \textbf{78.04} & 79.63 & \textbf{61.27} & \textbf{80.69} & \textbf{88.59} & \textbf{76.28} \\
			\Xhline{2\arrayrulewidth}
		\end{tabular}}}
		\vspace{.1em}
		\caption{\small Accuracy on CLEVR. CI, QA, CA stand for Count Integer, Query Attribute and Compare Attribute respectively. The top half uses ResNet-152 features and the bottom half uses ResNet-101 features. Our best model uses the same visual feature as \cite{johnson2016clevr}. }\label{tb:clevr}
		\vspace{-2.2em}
	\end{center}
\end{table}

\begin{table}
\begin{center}
\makebox[\linewidth]{\resizebox{0.68\linewidth}{!}{%
  \begin{tabular}{lccccc}
  \Xhline{2\arrayrulewidth}
  Model & All &  Y/N & No.  & Other \\
  \hline
  MCB\cite{fukui2016multimodal} &  64.7 & 82.5 & 37.6 & 55.6 \\
  MLB\cite{kim2016hadamard} &  65.08 & 84.14 & 38.21 & 54.87 \\
  \hline
  MF-SIG-T1 & 65.90 & 84.22 & 39.51 & 56.22  \\
  MF-SIG-T2 & 65.89 & 84.21 & 39.57 & 56.20  \\
  MF-SIG-T3 & \textbf{66.00} & \textbf{84.33} & \textbf{39.34} & \textbf{56.37}  \\
  MF-SIG-T4 & 65.81 & 84.22 & 38.96 & 56.16  \\
  LBP-SIG-T1 & 65.93 & 84.31 & 39.27 & 56.26  \\
  LBP-SIG-T2 & 65.90 & 84.23 & 39.70 & 56.16  \\
  LBP-SIG-T3 & 65.81 & 84.05 & 39.76 & 56.12  \\
  LBP-SIG-T4 & 65.73 & 84.08 & 38.87 & 56.13 \\
  \Xhline{2\arrayrulewidth}
  MCB+VG\cite{fukui2016multimodal} & 65.4 & 82.3 & 37.2 & 57.4 \\
  MLB+VG\cite{kim2016hadamard}  & 65.84 & 83.87 & 37.87 & 56.76 \\
  \hline
  MF+VG &  \textbf{67.17} & \textbf{84.77}& \textbf{39.71} & \textbf{58.34} \\
  MF-SIG+VG &  \textbf{67.19} & \textbf{84.71}& \textbf{40.58} & \textbf{58.24} \\
  \Xhline{2\arrayrulewidth}
  \multicolumn{4}{l}{On \texttt{test-dev2017} of VQA2.0} \\
  \hline
  MF-SIG+VG & 64.73 & 81.29 & 42.99 & 55.55  \\
  \Xhline{2\arrayrulewidth}
\end{tabular}}}
\vspace{.1em}
\caption{Results of the Open Ended task on \texttt{test-dev}.}\label{tb:VQA}
\vspace{-1.5em}
\end{center}
\end{table}

\begin{table}
\begin{center}
	\makebox[\linewidth]{\resizebox{0.89\linewidth}{!}{%
  \begin{tabular}{lcccccc}
  \Xhline{2\arrayrulewidth}
  & \multicolumn{4}{c}{Open Ended} & MC \\
  \cmidrule{2-6}
  Single Model & All &  Y/N & No.  & Other & All \\
  \hline
  SMem\cite{xu2016ask} & 58.24 & 80.8 & 37.53  & 43.48 & - \\
  SAN\cite{yang2016stacked} &  58.85 & 79.11 & 36.41 & 46.42 & - \\
  D-NMN\cite{andreas2016learning} &  59.4 & 81.1 & 38.6 & 45.5 & - \\
  ACK\cite{wu2016ask} &  59.44 & 81.07 & 37.12 & 45.83 & - \\
  FDA\cite{ilievski2016focused} &  59.54 & 81.34 & 35.67 & 46.10 & 64.18 \\
  QRU\cite{li2016visual} &  60.76 & - & - & - & 65.43 \\
  HYBRID\cite{kafle2016answer} & 60.06 & 80.34 & 37.82 & 47.56 & - \\
  DMN+\cite{xiong2016dynamic} & 60.36 & 80.43 & 36.82 & 48.33 & - \\
  MRN\cite{kim2016multimodal} & 61.84 & 82.39 & 38.23 & 49.41 & 66.33 \\
  HieCoAtt\cite{lu2016hierarchical} & 62.06 & 79.95 & 38.22 & 51.95 & 66.07 \\
  RAU\cite{noh2016training} & 63.2 & 81.7 & 38.2 & 52.8 & 67.3 \\
  MLB\cite{kim2016hadamard} &  65.07 & 84.02 & 37.90 & 54.77 & 68.89 \\
  \hline
  MF-SIG-T3 & \textbf{65.88} & \textbf{84.42} & \textbf{38.94} & \textbf{55.89} & \textbf{70.33}  \\
  \Xhline{2\arrayrulewidth}
  Ensemble Model\\
  \hline
  MCB\cite{fukui2016multimodal} & 66.47 & 83.24 & 39.47 & 58.00 & 70.10 \\
  MLB\cite{kim2016hadamard}  & 66.89 & 84.61 & 39.07 & 57.79 & 70.29  \\
  \hline
  Ours &  \textbf{68.14} & \textbf{85.41} & \textbf{40.99 } & \textbf{59.27} & \textbf{72.08} \\
  Ours \texttt{test2017} & 65.84 & 81.85 & 43.64 & 57.07 & - \\
  \Xhline{2\arrayrulewidth}
\end{tabular}}}
\vspace{.1em}
\caption{Results of the Open Ended and Multiple Choice tasks on \texttt{test}. We compare the accuracy of  single models (without augmentation) and ensemble models with published methods.}\label{tb:VQA_test}
\vspace{-1.8em}
\end{center}
\end{table}

\subsubsection{On the VQA dataset}
Since we have found MF-SIG and LBP-SIG are the best on CLEVR, in this part, we mainly compare the two models with different $T$. Notice now the total number of glimpses is the same as MCB \cite{fukui2016multimodal} and MLB \cite{kim2016hadamard}, and both of them use \texttt{res5c} features and better feature pooling methods. The optimal choice in these experiments is MF-SIG-T3, which is 0.92\% higher in overall accuracy than the previous best method \cite{kim2016hadamard}, and outperforms previous methods on all 3 general categories of questions. We then use external data from Visual Genome to train MF-SIG-T3 and MF-T3, in which MF-SIG surpassed MLB under the same condition by 1.35\%. The accuracy boost of our model is higher than MCB and MLB, showing that our model has higher capacity. The LBP models, which performs better than MF layers on CLEVR, turns out to be worse on this dataset, and $T=1$ is the optimal choice for LBP. We also find the single MF attention model, which should not be as powerful as MF-SIG, achieved 67.17\% accuracy with augmentation. These might be caused by the bias of the current VQA dataset \cite{antol2015vqa}, where there are questions with fixed answers across all involved images. We also show the results on \texttt{test}, as shown in Table \ref{tb:VQA_test}. Our model is the best among published methods without external data. With an ensemble of 3 MF-T3 and 4 MF-SIG-T3 models, we achieve 68.18\% accuracy on \texttt{test}, 1.25\% higher than best published ensemble model on the Open Ended task. By the date of submission, we rank the second on the leaderboard of Open Ended task and the first on that of the Multiple Choice task. The champion on Open Ended has an accuracy of 69.94\% but the method is not published. We have also recorded our model's performance on the \texttt{test-dev2017} and \texttt{test2017} of VQA2.0 in Table \ref{tb:VQA} and \ref{tb:VQA_test}. Accuracy on \texttt{test2017} is achieved with 8 snapshots from 4 models with different learning rates.
\vspace{-1.0em}
\subsubsection{Qualitative Results}
\begin{figure}
  \centering
  \includegraphics[width=\linewidth]{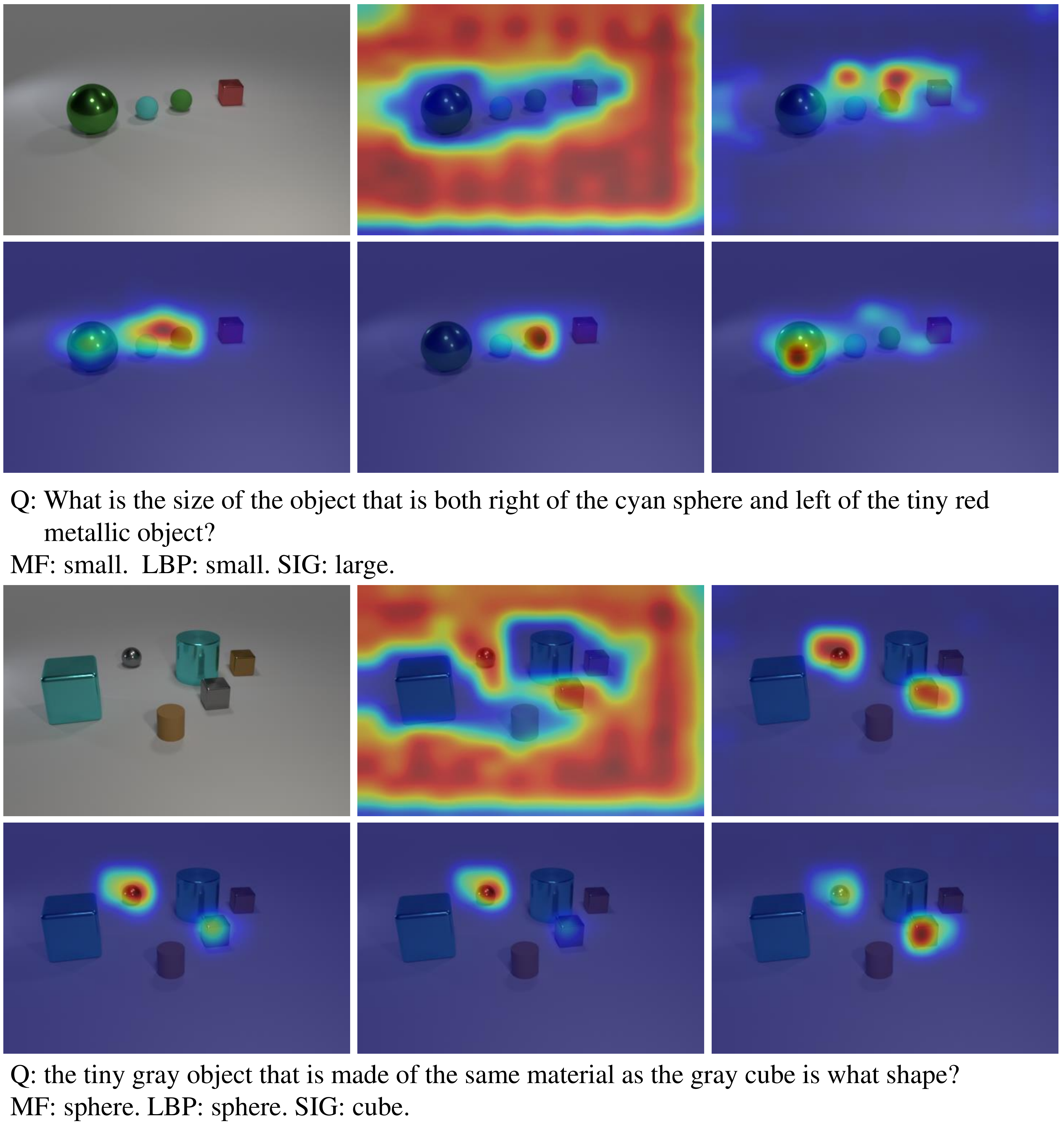}
  \caption{Two instances of different attentions on CLEVR, where the SIG model gives wrong answers but MF-SIG and LBP-SIG both give the correct answer. For each instance, from left to right, the first row to the second row, the images are: input image, $b^{(0)}$ of MF-SIG, $b^{(3)}$ of MF-SIG, $\psi_i(z_i)$ of LBP-SIG, $b$ of LBP-SIG, attention of SIG. Notations are the same as in Fig. \ref{fig:shapes_attention}. Best viewed in color.}\label{fig:clevr_qua}
  \vspace{-1.3em}
\end{figure}
We demonstrate some attention maps on CLEVR and the VQA dataset to analyze the behavior of the proposed models. Fig. \ref{fig:clevr_qua} shows 2 instances where the SIG model failed but both MF and LBP succeeded. We find the MF-SIG model has learned interesting patterns where its attention often covers the background surrounding the target initially, but converges to the target after iterative inference. This phenomenon almost never happens with the LBP-SIG model, which usually has better initializations that contained the target region. The shortcoming of the unstructured SIG model is also exposed in the 2 instances, where it tends to get stuck with the key nouns of the question.
\begin{figure}
  \centering
  \includegraphics[width=\linewidth]{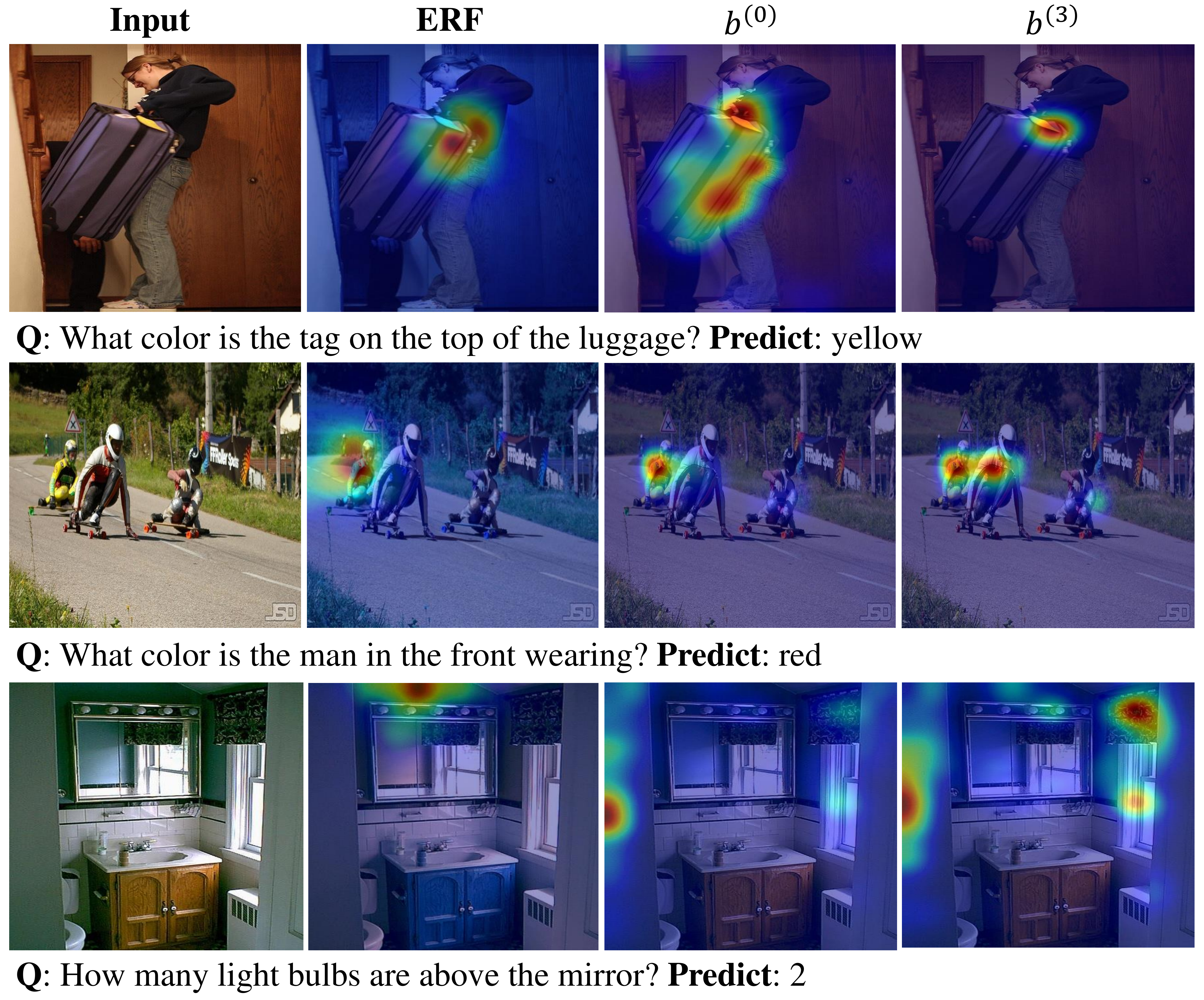}
  \caption{Some instances in the VQA dataset. The ERFs locate at the target region in row 1 and 3, and at at initial attention in row 2. Best viewed in color.}\label{fig:qualitative}
  \vspace{-1.3em}
\end{figure}
Fig. \ref{fig:qualitative} demonstrates 3 instances of the MF-SIG model together with the effective receptive field. The model gives 2 correct answers for the first 2 instances and 1 wrong answer for the last instance. In the first instance, the ERF at the target should be enough to encode the relations. The initial attention involves some extra areas due to the key word ``luggage", but it manages to converge to the most relevant region. In the second instance, the initial attention is wrong, as we can see the ERF at the initial attention does not overlap with the target, but with the help of MF, the final attention captures the relation ``in the front" and gives an acceptable answer. In the third instance, the ERF at the target region is very weak on the keyword ``bulb", which means the feature vector does not encode this concept, probably due to the size of the bulb. The model fails to attend to the right region and gives a popular answer ``2" (3rd most popular answer on the VQA dataset) according to the type of the question.

\section{Conclusion}
In this paper, we propose a novel structured visual attention mechanism for VQA, which models attention with binary latent variables and a grid-structured CRF over these variables. Inference in the CRF is implemented as recurrent layers in neural networks. Experimental results demonstrate that the proposed method is capable of capturing the semantic structure of the image in accordance with the question, which alleviates the problem of unstructured attention that captures only the key nouns in the questions. As a result, our method achieves state-of-the-art accuracy on three challenging datasets. Although structured visual attention does not solve all problems in VQA, we argue that it should be an indispensable module for VQA in the future.

{\small
\bibliographystyle{ieee}
\bibliography{egbib}

\begin{thebibliography}{10}\itemsep=-1pt

\bibitem{andreas2016learning}
J.~Andreas, M.~Rohrbach, T.~Darrell, and D.~Klein.
\newblock Learning to compose neural networks for question answering.
\newblock {\em NAACL}, 2016.

\bibitem{andreas2016neural}
J.~Andreas, M.~Rohrbach, T.~Darrell, and D.~Klein.
\newblock Neural module networks.
\newblock In {\em CVPR}, 2016.

\bibitem{antol2015vqa}
S.~Antol, A.~Agrawal, J.~Lu, M.~Mitchell, D.~Batra, C.~Lawrence~Zitnick, and
  D.~Parikh.
\newblock {VQA}: Visual question answering.
\newblock In {\em ICCV}, 2015.

\bibitem{chen2015learning}
L.-C. Chen, A.~G. Schwing, A.~L. Yuille, and R.~Urtasun.
\newblock Learning deep structured models.
\newblock In {\em ICML}, 2015.

\bibitem{chen2015mxnet}
T.~Chen, M.~Li, Y.~Li, M.~Lin, N.~Wang, M.~Wang, T.~Xiao, B.~Xu, C.~Zhang, and
  Z.~Zhang.
\newblock Mxnet: A flexible and efficient machine learning library for
  heterogeneous distributed systems.
\newblock {\em arXiv preprint arXiv:1512.01274}, 2015.

\bibitem{das2016human}
A.~Das, H.~Agrawal, C.~L. Zitnick, D.~Parikh, and D.~Batra.
\newblock Human attention in visual question answering: Do humans and deep
  networks look at the same regions?
\newblock {\em EMNLP}, 2016.

\bibitem{do2010neural}
T.-M.-T. Do and T.~Artieres.
\newblock Neural conditional random fields.
\newblock In {\em AISTATS}, 2010.

\bibitem{fukui2016multimodal}
A.~Fukui, D.~H. Park, D.~Yang, A.~Rohrbach, T.~Darrell, and M.~Rohrbach.
\newblock Multimodal compact bilinear pooling for visual question answering and
  visual grounding.
\newblock {\em EMNLP}, 2016.

\bibitem{gal2016theoretically}
Y.~Gal and Z.~Ghahramani.
\newblock A theoretically grounded application of dropout in recurrent neural
  networks.
\newblock In {\em NIPS}, 2016.

\bibitem{he2016deep}
K.~He, X.~Zhang, S.~Ren, and J.~Sun.
\newblock Deep residual learning for image recognition.
\newblock In {\em CVPR}, 2016.

\bibitem{ilievski2016focused}
I.~Ilievski, S.~Yan, and J.~Feng.
\newblock A focused dynamic attention model for visual question answering.
\newblock {\em arXiv preprint arXiv:1604.01485}, 2016.

\bibitem{jaderberg2014deep}
M.~Jaderberg, K.~Simonyan, A.~Vedaldi, and A.~Zisserman.
\newblock Deep structured output learning for unconstrained text recognition.
\newblock {\em ICLR}, 2015.

\bibitem{johnson2016clevr}
J.~Johnson, B.~Hariharan, L.~van~der Maaten, L.~Fei-Fei, C.~L. Zitnick, and
  R.~Girshick.
\newblock Clevr: A diagnostic dataset for compositional language and elementary
  visual reasoning.
\newblock {\em CVPR}, 2017.

\bibitem{kafle2016answer}
K.~Kafle and C.~Kanan.
\newblock Answer-type prediction for visual question answering.
\newblock In {\em CVPR}, 2016.

\bibitem{kafle2016visual}
K.~Kafle and C.~Kanan.
\newblock Visual question answering: Datasets, algorithms, and future
  challenges.
\newblock {\em arXiv preprint arXiv:1610.01465}, 2016.

\bibitem{kim2016multimodal}
J.-H. Kim, S.-W. Lee, D.~Kwak, M.-O. Heo, J.~Kim, J.-W. Ha, and B.-T. Zhang.
\newblock Multimodal residual learning for visual {QA}.
\newblock In {\em NIPS}, 2016.

\bibitem{kim2016hadamard}
J.-H. Kim, K.-W. On, J.~Kim, J.-W. Ha, and B.-T. Zhang.
\newblock Hadamard product for low-rank bilinear pooling.
\newblock {\em ICLR}, 2017.

\bibitem{kim2017structured}
Y.~Kim, C.~Denton, L.~Hoang, and A.~M. Rush.
\newblock Structured attention networks.
\newblock {\em ICLR}, 2017.

\bibitem{kingma2014adam}
D.~Kingma and J.~Ba.
\newblock Adam: A method for stochastic optimization.
\newblock {\em ICLR}, 2015.

\bibitem{kiros2015skip}
R.~Kiros, Y.~Zhu, R.~R. Salakhutdinov, R.~Zemel, R.~Urtasun, A.~Torralba, and
  S.~Fidler.
\newblock Skip-thought vectors.
\newblock In {\em NIPS}, 2015.

\bibitem{krishna2016visual}
R.~Krishna, Y.~Zhu, O.~Groth, J.~Johnson, K.~Hata, J.~Kravitz, S.~Chen,
  Y.~Kalantidis, L.-J. Li, D.~A. Shamma, et~al.
\newblock Visual genome: Connecting language and vision using crowdsourced
  dense image annotations.
\newblock {\em IJCV}, 2016.

\bibitem{li2016visual}
R.~Li and J.~Jia.
\newblock Visual question answering with question representation update
  ({QRU}).
\newblock In {\em NIPS}, 2016.

\bibitem{lin2014microsoft}
T.-Y. Lin, M.~Maire, S.~Belongie, J.~Hays, P.~Perona, D.~Ramanan,
  P.~Doll{\'a}r, and C.~L. Zitnick.
\newblock Microsoft {COCO}: Common objects in context.
\newblock In {\em ECCV}, 2014.

\bibitem{lu2016hierarchical}
J.~Lu, J.~Yang, D.~Batra, and D.~Parikh.
\newblock Hierarchical question-image co-attention for visual question
  answering.
\newblock In {\em NIPS}, 2016.

\bibitem{luo2016understanding}
W.~Luo, Y.~Li, R.~Urtasun, and R.~Zemel.
\newblock Understanding the effective receptive field in deep convolutional
  neural networks.
\newblock In {\em NIPS}, 2016.

\bibitem{malinowski2014multi}
M.~Malinowski and M.~Fritz.
\newblock A multi-world approach to question answering about real-world scenes
  based on uncertain input.
\newblock In {\em NIPS}, 2014.

\bibitem{nam2016dual}
H.~Nam, J.-W. Ha, and J.~Kim.
\newblock Dual attention networks for multimodal reasoning and matching.
\newblock {\em arXiv preprint arXiv:1611.00471}, 2016.

\bibitem{noh2016training}
H.~Noh and B.~Han.
\newblock Training recurrent answering units with joint loss minimization for
  {VQA}.
\newblock {\em arXiv preprint arXiv:1606.03647}, 2016.

\bibitem{nowozin2011structured}
S.~Nowozin, C.~H. Lampert, et~al.
\newblock Structured learning and prediction in computer vision.
\newblock {\em Foundations and Trends{\textregistered} in Computer Graphics and
  Vision}, 6(3--4):185--365, 2011.

\bibitem{peng2009conditional}
J.~Peng, L.~Bo, and J.~Xu.
\newblock Conditional neural fields.
\newblock In {\em NIPS}, 2009.

\bibitem{shih2016look}
K.~J. Shih, S.~Singh, and D.~Hoiem.
\newblock Where to look: Focus regions for visual question answering.
\newblock In {\em CVPR}, 2016.

\bibitem{shimony1994finding}
S.~E. Shimony.
\newblock Finding maps for belief networks is np-hard.
\newblock {\em Artificial Intelligence}, 68(2):399--410, 1994.

\bibitem{srivastava2014dropout}
N.~Srivastava, G.~E. Hinton, A.~Krizhevsky, I.~Sutskever, and R.~Salakhutdinov.
\newblock Dropout: a simple way to prevent neural networks from overfitting.
\newblock {\em JMLR}, 2014.

\bibitem{tu2014joint}
K.~Tu, M.~Meng, M.~W. Lee, T.~E. Choe, and S.-C. Zhu.
\newblock Joint video and text parsing for understanding events and answering
  queries.
\newblock {\em IEEE MultiMedia}, 2014.

\bibitem{weiss2001comparing}
Y.~Weiss.
\newblock Comparing the mean field method and belief propagation for
  approximate inference in mrfs.
\newblock {\em Advanced Mean Field Methods—Theory and Practice}, pages
  229--240, 2001.

\bibitem{wu2016ask}
Q.~Wu, P.~Wang, C.~Shen, A.~Dick, and A.~van~den Hengel.
\newblock Ask me anything: Free-form visual question answering based on
  knowledge from external sources.
\newblock In {\em CVPR}, 2016.

\bibitem{xiong2016dynamic}
C.~Xiong, S.~Merity, and R.~Socher.
\newblock Dynamic memory networks for visual and textual question answering.
\newblock {\em ICML}, 1603, 2016.

\bibitem{xu2016ask}
H.~Xu and K.~Saenko.
\newblock Ask, attend and answer: Exploring question-guided spatial attention
  for visual question answering.
\newblock In {\em ECCV}, 2016.

\bibitem{xu2015show}
K.~Xu, J.~Ba, R.~Kiros, K.~Cho, A.~C. Courville, R.~Salakhutdinov, R.~S. Zemel,
  and Y.~Bengio.
\newblock Show, attend and tell: Neural image caption generation with visual
  attention.
\newblock In {\em ICML}, 2015.

\bibitem{yang2016stacked}
Z.~Yang, X.~He, J.~Gao, L.~Deng, and A.~Smola.
\newblock Stacked attention networks for image question answering.
\newblock In {\em CVPR}, 2016.

\bibitem{yeh2008photo}
T.~Yeh, J.~J. Lee, and T.~Darrell.
\newblock Photo-based question answering.
\newblock In {\em ACMMM}, 2008.

\bibitem{zheng2015conditional}
S.~Zheng, S.~Jayasumana, B.~Romera-Paredes, V.~Vineet, Z.~Su, D.~Du, C.~Huang,
  and P.~H. Torr.
\newblock Conditional random fields as recurrent neural networks.
\newblock In {\em ICCV}, 2015.

\bibitem{zitnick2014edge}
C.~L. Zitnick and P.~Doll{\'a}r.
\newblock Edge boxes: Locating object proposals from edges.
\newblock In {\em ECCV}, 2014.

\end{thebibliography}
}

\end{document}